\documentclass{article}
\usepackage{ijcai24}

\usepackage{times}
\usepackage{soul}
\usepackage{url}
\usepackage[hidelinks]{hyperref}
\usepackage[utf8]{inputenc}
\usepackage[small]{caption}
\usepackage{graphicx}
\usepackage{amsmath}
\usepackage{amsthm}
\usepackage{booktabs}
\usepackage{algorithm}
\usepackage{algorithmic}
\usepackage[switch]{lineno}
\usepackage{bbding}
\usepackage{pifont}
\usepackage{wasysym}
\usepackage{utfsym}
\usepackage{fontawesome}
\usepackage{enumitem}
\usepackage{subfig}
\usepackage{multirow}
\usepackage{mathtools}
\usepackage{amssymb}
\usepackage{amsfonts}

\usepackage{float}
\usepackage{adjustbox}
\usepackage{bm}

\title{Score-CDM: Score-Weighted Convolutional Diffusion Model \\for Multivariate Time Series Imputation}

\author{
Shunyang Zhang$^{1}$\
\and
Senzhang Wang$^{1}$\and
Hao Miao$^{2}$\and
Hao Chen$^3$ \\
Changjun Fan$^4$ \And
Jian Zhang$^1$  \\
\affiliations
$^1$Central South University \\
$^2$Aalborg University  \\
$^3$The Hong Kong Polytechnic University \\
$^4$National University of Defense Technology\\
\emails
\{224712166, szwang\}@csu.edu.cn,
haom@cs.aau.dk,
sundaychenhao@gmail.com,
fanchangjun@nudt.edu.cn,
jianzhang@csu.edu.cn
}

\begin{document}

\maketitle

\begin{abstract}

Multivariant time series (MTS) data are usually incomplete in real scenarios, and imputing the incomplete MTS is practically important to facilitate various time series mining tasks. Recently, diffusion model-based MTS imputation methods have achieved promising results by utilizing CNN or attention mechanisms for temporal feature learning. However, it is hard to adaptively trade off the diverse effects of local and global temporal features by simply combining CNN and attention. To address this issue, we propose a Score-weighted Convolutional Diffusion Model (Score-CDM for short), whose backbone consists of a Score-weighted Convolution Module (SCM) and an Adaptive Reception Module (ARM). SCM adopts a score map to capture the global temporal features in the time domain, while ARM uses a Spectral2Time Window Block (S2TWB) to convolve the local time series data in the spectral domain. Benefiting from the time convolution properties of Fast Fourier Transformation, ARM can adaptively change the receptive field of the score map, and thus effectively balance the local and global temporal features. We conduct extensive evaluations on three real MTS datasets of different domains, and the result verifies the effectiveness of the proposed Score-CDM.

\end{abstract}

\section{Introduction}




The perpetual integration of sensing technologies results in the generation of increasingly voluminous time series data, contributing to various practical applications such as urban planning \cite{urbanplan}, city renewal \cite{timeas}, and traffic management \cite{wang2020deep,miao2024unified,astcmcn,simplets,mdtp},. Nevertheless, in practical scenarios, the time series data are usually incomplete due to various issues including sensor failure and communication errors. To this end, multivariate time series (MTS) imputation has attracted rising research interest in both academic and industrial communities in recent years, finding applications across diverse domains such as finance, healthcare, and industrial manufacturing \cite{timeas}.

Traditionally, statistical-based machine learning methods are widely adopted for MTS imputation, such as ARIMA \cite{arima} and KNN \cite{knn}. These methods typically can capture the linear properties of time series, but may not be effective to model the complex and nonlinear temporal correlations. Recent progress in deep learning has brought about more effective MTS imputation techniques, including RNN, CNN, and Attention. RNNs continuously update the hidden states to capture the temporal information for time series data. Some RNN-based methods \cite{grin,astcmcn,li2022fine} adopt attention to further consider the temporal feature correlations. A major issue of RNN-based methods is the accumulation of errors  \cite{PriSTI} that can result in inaccurate imputations. Attention-based models~\cite{spin} can effectively model the long-term temporal features \cite{inception}, but may overlook the local and short-term temporal correlation. Although TCN can effectively capture the long-term dependencies of time series due to its larger receptive field \cite{scinet}, its temporal feature learning capacity is still largely limited by the size of the receptive field.




Recently, diffusion models (DM) \cite{denoising}, recognized by their powerful generative learning capability and great success in data generation tasks, have also been applied to impute MTS and achieved SOTA performance \cite{csdi,PriSTI}. These models tackle the data imputation task by creating conditional guidance, aiming to bring the diffusion process and backward process to an accurate outcome. In simpler terms, they create a process that learns a map from the ground truth to noise and another map to reconstruct data from noise \cite{score}. The flexibility of DM in neural network architecture allows it to incorporate different deep learning models (e.g. CNN and attention) as its denoising function \cite{csdi}.

Despite their effectiveness, existing diffusion models have overlooked the design of a suitable denoising function that can effectively capture the global and local temporal features for time series imputation \cite{PriSTI}. Directing adopting attention mechanism as the denoising function can effectively capture global temporal features (e.g., weekly traffic flow patterns), but the local temporal features (e.g., hourly traffic flow features) may not be effectively learned \cite{inception}. TCN can model larger and structured receptive fields by combining structured transformation with CNN. However, it still lacks of a broader receptive field compared to attention mechanisms. Approaches like attention-free transformer \cite{aft} attempt to integrate CNN to capture the local correlations. However, this type of method handles the local and global temporal features separately, and thus is hard to adaptively balance the diverse effects of the two types of features. 

In this paper, we propose a \textbf{Score}-Weighted \textbf{C}onvolutional \textbf{D}iffusion \textbf{M}odel named Score-CDM to effectively and adaptively learn the local and global time series features for MTS imputation. Specifically, Score-CDM contains a Score-weighted Convolution Module (SCM) and an Adaptive Reception module (ARM). SCM can generate a globally attentive convolution kernel, and ARM can construct a time window that regulates the receptive field of this kernel. We derive a globally attentive convolution operation by multiplying this kernel with the elements within the receptive field. To delve deeper, we propose a  Spectral2time Window Block (S2TWB) in ARM, which can adaptively change the receptive field in the spectral domain by using the Fast Fourier Transform (FFT). By leveraging the convolution properties of FFT, S2TWB establishes a flexible receptive field to achieve a balance between the global and local temporal features by learning to weight each time step in the time window. SCM and ARM together work as the denoising function of Score-CDM for more effective MTS imputation. We conduct extensive evaluations on three real-world MTS datasets. The results show the superior performance of our proposed method by comparison with current SOTA models. We summarize our contributions as follows.

\begin{itemize}

\item We introduce Score-CDM, a score-weighted convolutional diffusion model whose denoising function contains the novel SCM and ARM. Score-CDM can more effectively capture the global and local temporal features for MTS imputation.

\item  A spectral2time window block S2TWB is proposed to adaptively change the receptive field of the kernel generated by SCM on the spectral domain by adopting FFT. Due to the time convolution properties of FFT, the extracted receptive field is structured and flexible (e.g. continuous or discretized).

\item We conduct extensive evaluations on three real-world MTS datasets of different domains. The result verifies the superior performance of our proposal, and also demonstrates the effectiveness of the two proposed modules SCM and ARM in temporal feature learning. 

\end{itemize}

\section{Related Work}
Multivariate time series imputation attracts increasing interest due to the increasing availability of time series data and rich applications, such as statistical methods \cite{MF} and deep models \cite{s4d}. In the early stage, time series imputation methods are mostly based on statistical models, such as Matrix Factorization (MF) \cite{MF}, and Multiple Imputation using Chained Equations. However, their linear properties make them hard to capture the dynamic features of time series. Recently, various deep learning-based methods are proposed to address time series imputation \cite{RNN1,RNN2,RNN3}. BRITS utilized a simple linear regression layer to incorporate spatial information and adopted bidirectional RNNs architecture for time series imputation \cite{BRITS}. SAITS introduced self-attention to capture the global temporal relations of time series \cite{saits}. SPIN adopted a joint attention that combined spatial and temporal attention to model information exchange between time series variants ~\cite{spin}. 

Motivated by the great success of generative models, multiple generative model based MTS imputation methods are also proposed and achieved SOTA performance. GAINFilling used GAN models to generate sequences by matching the underlying data distribution \cite{gainfilling}. Conditional Score-based Diffusion models for Imputation (CSDI) is a paradigmatic example of applying the diffusion models in MTS imputation \cite{csdi}. CSDI presented a novel time series imputation method that leveraged score-based diffusion models. Following CSDI, Structured State Space Diffusion (SSSD)  integrated conditional diffusion models and structured state-space models to particularly capture long-term dependencies in time series \cite{sssd}. PriSTI applied spatial information to guide the generation of the missing time series values \cite{PriSTI}. Unlike CSDI and SSSD, TimeDiff \cite{NARDIff} introduced additional inductive bias in the conditioning module to achieve long-time series forecasting. Diffusion model based methods generally perform well in time series imputation.  However, existing diffusion model based methods still suffer from the issue of lacking sufficient temporal and global temporal feature learning capacity because their backbone denoising function directly adopts attention or CNN models. How to design a new denoising function that is more suitable to deal with the MTS data is not well studied.

\newtheorem{problemdefinition}{Problem Definition}

\section{Preliminary and Problem Definition}

In this section, we will first define the studied problem, and then briefly introduce Fourier transformation and diffusion probabilistic models. 
\begin{problemdefinition}
    Given the incomplete multivariate time series ${X}\in \mathbb{R}^{N \times C \times L}$ with some missing values, we aim to build a model $\epsilon_\theta$ to impute ${X} \in \mathbb{R}^{N \times C \times L}$ and obtain the complete data $\widetilde{X}$, where $N$ is the number of time series variables, $C$ is the number of channels and $L$ is the length of the time series.
\end{problemdefinition}

\paragraph{\textbf{Fourier Operator} \label{sec:FO}}
We define $\mathcal{S}=\mathcal{F}(\kappa)\in\mathbb{C}^{N\times C \times L}$ as a Fourier Operator (FO), where $\mathcal{F}$ denotes Discrete Fourier Transform (DFT). According to the convolution theorem (see Appendix), we can write the multiplication between $\mathcal{F}({X})$ and $\mathcal{S}$ in Fourier space as follows,
\begin{equation}\label{equ:fgso}
\begin{aligned}
     \mathcal{F}({X})\mathcal{F}(\kappa) & = \mathcal{F}((X*\kappa)[i]) \\
    & =\mathcal{F}(\sum_{j=1}^n {{X}}[j]\kappa[i-j]) \\
    & =\mathcal{F}(\sum_{j=1}^n {{X}}[j]\kappa[i,j]), \quad\quad   \forall i \in [1,...,n]
\end{aligned}
\end{equation}
where $(X*\kappa)[i]$ denotes the convolution of $X$ and $\kappa$. As defined $\kappa[i,j]= W$ ($W \in R^{n\times n} $), it yields $\sum_{j=1}^n {X}[j]\kappa[i,j]=\sum_{j=1}^n {X}[j]W=XW$. Accordingly, we can get the convolution equation as follows,
\begin{equation}\label{equ:gcn_unit}
    \mathcal{F}(X)\mathcal{S} = \mathcal{F}(XW).
\end{equation}
From Eq.\ref{equ:gcn_unit}, one can observe that performing the multiplication between $\mathcal{F}(X)$ and $\mathcal{S}$ in Fourier space corresponds to a time shift operation in Eq.\ref{equ:fgso}  (i.e., a temporal convolution) in the time domain. Since the multiplication in Fourier space has much lower complexity ($\mathcal{O}(t \log t)$) than the above shift operations ($\mathcal{O}(t^2)$) in the time domain, we adopt FFT to make a more efficient convolution on the time domain.

\paragraph{\textbf{Diffusion process and reverse process.} \label{subsec:diff}}
The \textit{diffusion process} for MTS imputation adds Gaussian noise into the original data, which can be formalized as follows,
\begin{equation}
\begin{aligned}
&q(\widetilde{X}^{1:T}|\widetilde{X}^{0})=\prod_{t=1}^T q(\widetilde{X}^{t}|\widetilde{X}^{t-1}), \\
&q(\widetilde{X}^{t}|\widetilde{X}^{t-1})=\mathcal{N}(\widetilde{X}^{t}; \sqrt{1-\beta_{t}}\widetilde{X}^{t-1}, \beta_t \bm{I}),
\end{aligned}
\end{equation}
where $\beta_t$ is a small constant hyperparameter that controls the variance of the added noise.
$\widetilde{X}^t$ is sampled by $\widetilde{X}^t=\sqrt{\bar{\alpha}_t}\widetilde{X}^0+\sqrt{1-\bar{\alpha}_t}\epsilon$, where $\alpha_t=1-\beta_t$, $\bar{\alpha}_t=\prod_{i=1}^t\alpha_i$, and $\epsilon$ is the sampled standard Gaussian noise. When $T$ is large enough, $q(\widetilde{X}^T|\widetilde{X}^0)$ is close to the standard normal distribution.

The \textit{reverse process} for MTS imputation gradually converts random noise to the missing values with spatiotemporal consistency based on conditional information. In this work, the reverse process is conditioned on the interpolated conditional information $\mathcal{X}$ (conditional guidance) that enhances the observed values, which can be formalized as follows,
\begin{equation}\label{eq:reverse_process}
\begin{aligned}
    & p_{\theta}(\widetilde{X}^{0:T-1}|\widetilde{X}^{T}, \mathcal{X})=\prod_{t=1}^T p_{\theta}(\widetilde{X}^{t-1}|\widetilde{X}^{t}, \mathcal{X}), \\
    & p_{\theta}(\widetilde{X}^{t-1}|\widetilde{X}^{t}, \mathcal{X})=\mathcal{N}(\widetilde{X}^{t-1}; \mu_{\theta}(\widetilde{X}^{t}, \mathcal{X}, t), \sigma_t^2 \bm{I}).
\end{aligned}
\end{equation}

\section{Methodology}

\begin{figure}[!t]
\centering
\includegraphics[width=\linewidth]{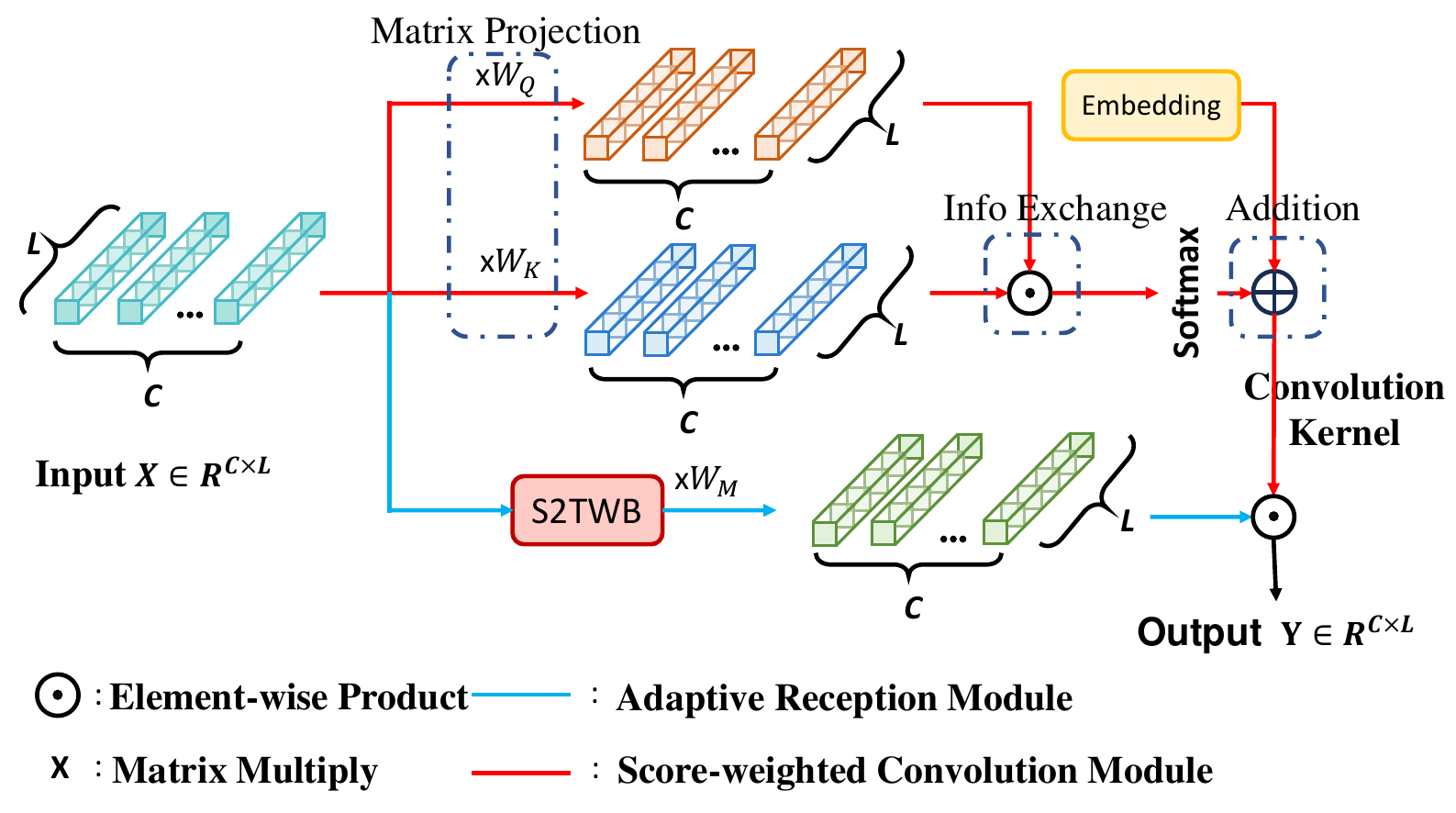}
\caption{The framework of the Score-CDM denoising function.  }
\label{fig1}
\end{figure}

The schematic representation of the denoising function of Score-CDM in the diffusion process is depicted in Figure \ref{fig1}. The designed denoising function mainly contains a Score-weighted convolution module (SCM) and a Adaptive Reception Module. The SCM undergoes two key steps: matrix projection and information exchange (the upper red line part). ARM undergoes one step: receptive field generation (the lower blue line part). The matrix projection involves the operation of multiplying raw data $X_i \in R^{C \times L}$ with two learnable matrices $W_K$ and $W_Q$, whose function is similar to the attention mechanism. In the information exchange phase, the element-wise product of $Q$ and $K$ is computed, facilitating the comprehensive multiplication of time series elements. This process enables the learning of a globally attentive score map through the application of Softmax and an additional embedding. For ARM, it contains a S2TWB block whose detailed illustration is given in Figure \ref{fig2}. As indicated by the blue line in Figure{ \ref{fig1}}, the raw data undergoes processing through the S2TWB block and matrix $W_M$, resulting in data aggregation within a specified time window. The time window (receptive field) from ARN and the score map from SCM are then multiplied using the element-wise product. Essentially, the score map acts as a convolution kernel to convolve the elements in the time window (receptive field) for this kernel. Next, we will provide a detailed exploration of these components.



\subsection{Score-weighted Convolution Module}

\paragraph{Matrix Projection.\label{time mix}} This step aims to aggregate the information of each time step in a time series. $X$ is the input multivariant time series, and $W$ is the injection matrix. $X_i$ is the $i$-th row of $X$ and $W_j$ is the $j$-th column of $W$.
\begin{equation}
\label{eq:time-mixing}
\begin{aligned}
    &Q_{i,j} = X_iW^Q_j, K_{i,j} = X_iW^K_j, M_{i,j} = X_iW^M_j\\
    &X \in \mathbb{R}^{C \times L}, W \in \mathbb{R}^{L \times L},
\end{aligned}
\end{equation}
We learn $Q$ and $K$ in a similar way as attention. After this operation, we send $Q$ and $K$ to the next nonlinear element-wise mixing step for information exchange.

\paragraph{Information Exchange.\label{ele mix}} In this step, the product of element pairs allows for a thorough computation on the features of $X_{i,j}$. To consider the global correlation between element pairs, the output of information exchange undergoes a Softmax layer to weight the importance of all $X_{i,j}, j=1,...,C$. However, traditional dot product operation considers all time series of $X_i$, and the combination with Softmax leads to redundant computation on element pairs. To address this issue, we generate a series of convolutional kernels whose size is equal to $Q$ as follows

\begin{equation}
\label{eq:element-mixing}
\begin{aligned}
    &Q_{i,j} \odot K_{i,j} = (X_iW^Q_j)(X_iW^K_j),
\end{aligned}
\end{equation}
based on which we can further derive 
\begin{equation}
\label{eq:element}
\begin{aligned}
    \sum X_iW_j \times \sum X_iW_j \rightarrow \sum_k \sum_m X_{i,k}X_{i,m} W W.
\end{aligned}
\end{equation}
For the attention mechanism, its computation of time series is equal to our model but greater on channel aspect (different channel of time series) in element view (more details will presented in Appendix \ref{sec:appendix}):
\begin{equation}
\label{eq:element-mixing}
\begin{aligned}
   \sum_k (\sum X_iW_k \times \sum X_kW_j) \rightarrow \sum_k \sum_p \sum_q X_{i,p}X_{k,q} W.
\end{aligned}
\end{equation}
Then, through the Softmax function, we gain the weighted kernel (score map) as follows,
\begin{equation}
\label{eq:adding}
\begin{aligned}
    Kernel = softmax(Q_{i,j} \odot K_{i,j}).
\end{aligned}
\end{equation}
Finally, we add an embedding to the kernel as a random shift operation similar to  Attention-Free Transformer \cite{aft},
\begin{equation}
\label{eq:e}
\begin{aligned}
    Kernel = Kernel + Embedding.
\end{aligned}
\end{equation}
The embedding is initialized in random and it is similar to an attention weights shift as in \cite{aft}.
\subsection{Adaptive Reception Module}

\paragraph{Spectral2Time Window Block. \label{st}}

This module uses a self-attention kernel named \textbf{S}pectral\textbf{2T}ime \textbf{W}indow \textbf{B}lock (S2TWB) to convolve the time series based on the convolution theorem in Eq. \ref{equ:fgso} as follows
\begin{equation}
\mathcal{F}(K \ast X) = \mathcal{F}(K) \cdot \mathcal{F}(X),
\end{equation}
where $\ast$ is a convolution operation and $\cdot$ is a multiply operation. $\mathcal{F}$ is the Fast Fourier Transform.
We aim to generate a series of kernels as $\mathcal{K}_{\theta_{i}}$ and aggregate them together to generate the kernel 
$ \mathcal{K}_{\theta}$ as follows
\begin{equation}
	\begin{split}\label{equ:spectraloperator}
 \mathcal{K}_{\theta}=\sum_{i=1}^{L}w_{i}\mathcal{K}_{\theta_{i}},\\
	\end{split}
\end{equation}
where $L$ is the time series length and $\mathcal{K}_{\theta_{i}}$ ($i=1$ to $L$) are basis operators with learnable parameters $\{w_{i}\}_{i=1}^{L}$. Here we use $sin()$ function to present the basic operator, and we have
\begin{equation}
	\begin{split}\label{equ:basisdefinition}
        \mathcal{K}_{\theta_{i}}\big(X) &= \sin\big(iX).
	\end{split}
\end{equation}
Then the kernel $\mathcal{K}_{\theta}$ of S2TWB can be reformulated as follows
\begin{equation}
	\begin{split}\label{equ:specblock}
\mathcal{K}_{\theta}=\mathbf{w}_{\text{sin}}\begin{bmatrix}
\sin(x)\\
\vdots \\
\sin(ix)
\end{bmatrix}
\end{split}
\end{equation}
where $\mathbf{w}_{\text{sin}}[i]=w_{i}$. Finally, we apply FFT on kernel $\mathcal{K}_{\theta}$ and $X$ as follows,
\begin{equation}
\mathcal{K}_{\theta} \ast X = \mathcal{F}^{-1}(\mathcal{F}(\mathcal{K}_{\theta}) \cdot \mathcal{F}(X)).
\label{eq:sker}
\end{equation}
\begin{figure}[!t]
\centering
\includegraphics[width=\linewidth]{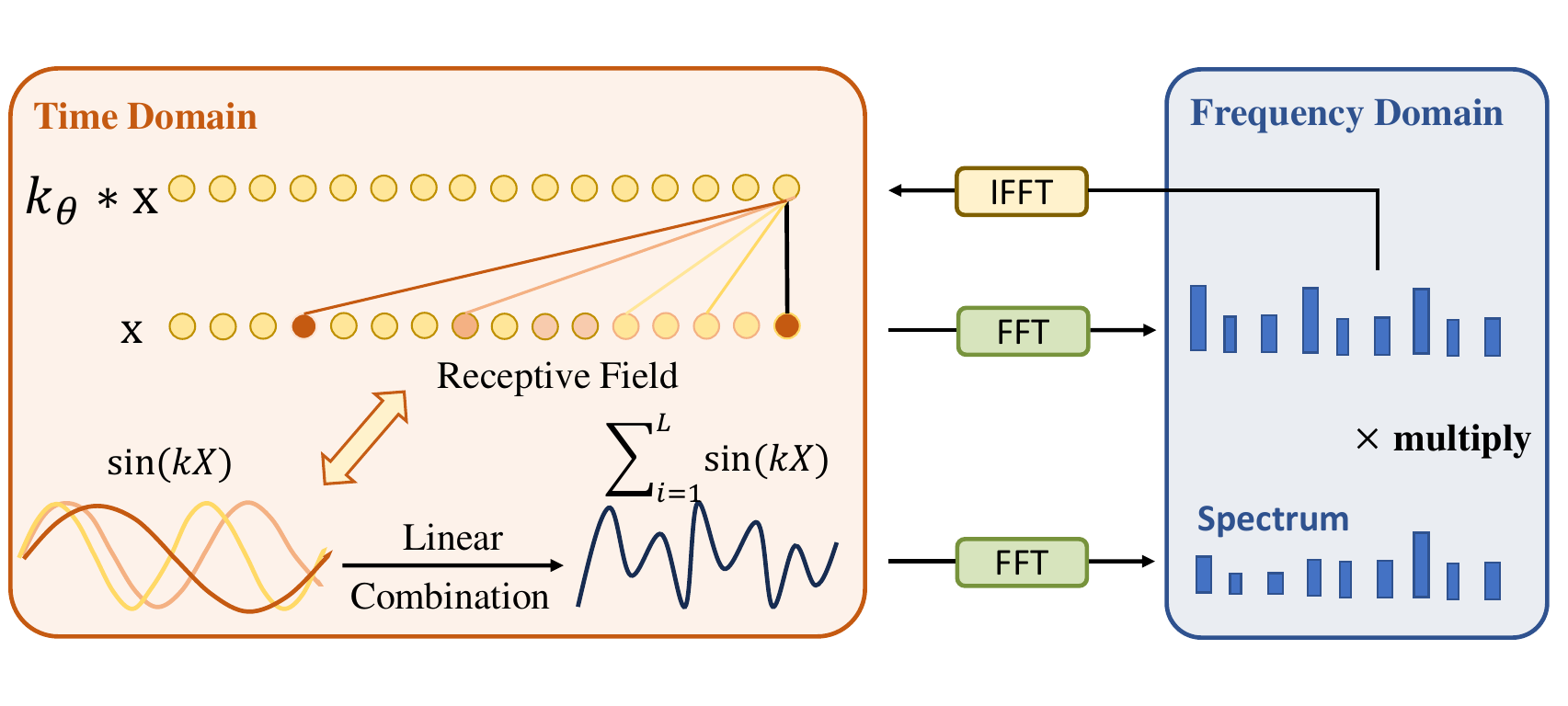}
\caption{Illustration of Spectral2Time Window Block}
\label{fig2}
\end{figure}

Next, we will demonstrate how to use FFT and kernel $\mathcal{K}_{\theta}$ to adaptively change the receptive field. FFT is a convolutional method for time series as mentioned in section \ref{sec:FO}. As shown in Figure \ref{fig2}, S2TWB uses FFT to project kernel and time series $X$ in the time domain to the spectral domain and multiply them, after which the kernel generated in the time domain convolves time series like the CNN kernel. Specifically, the kernel $\mathcal{K}_{\theta}$ is combined linearly with several waves as shown in the left side of Figure \ref{fig2}. Then the kernel allocates different attention weights to different positions of a time series based on the distance between itself and the target position in the time domain as shown in the left part of Figure \ref{fig2}. Convolution via FFT only relies on the relative position but not on the absolute position, which is more flexible.

\paragraph{Overall formulation.}
The overall mathematical presentation of the denoising function can be written as follows
\begin{equation}
\begin{aligned}
\label{eq:gcs-general}
  &Y = f(X); \\
  &Y_{c,t'} =  \frac{\mathcal{K}_{\theta} \ast((\exp(Q_{c,t'} \odot K_{c,t'}) + w_{c,t'}) \odot M_{c,t'})}{\sum_{t'=1}^T\exp(Q_{c,t'} \odot K_{c,t'})},
\end{aligned}
\end{equation}
where $Y$ is the output of the representation, $X$ is the input and $t_\tau$ is the structured receptive field. Then we learn a convolution operation as follows
\begin{equation}
\label{eq:correspond}
\mathcal{K}_{\theta (i,j)}Q_{i,j} = \mathcal{K}_{\theta (i,j)}X_{i,t_\tau}W^M_{j,t_\tau},
\end{equation}
which presents the $\mathcal{K}_{\theta (i,j)}$ convolves the $X_{i,t_\tau}$.


\subsection{Discussion}
\paragraph{Relationship with Convolution.} The convolutional kernel \(K\) projects onto the time series data \(X\), essentially performing a dot product between two vectors. In Score-CDM, the transformation matrix \(M\) projects a portion of the vector \(X_{i,t_0:t_1}\) onto unit elements, and the result is multiplied by a convolutional kernel with a size of \(1 \times 1\). This is equivalent to convolving \(X_{i,t_0:t_1}\) with a kernel of size \((t_1 - t_0) \times 1\).
\paragraph{Relationship with Transformer.}
For the attention mechanism, after time-mixing and element-mixing, the temporal values on the nodes are thoroughly blended, resulting in \( \sum X \times \sum X \rightarrow \sum_i \sum_j X_iX_j\). At this point, applying Softmax yields attention weights that span the entire time series.

We compare Score-CDM with four classic models including Attention Free Transformer, Transformer, TCN, and Dlinear (MLP-based Model) from four aspects, whether can be coded as attention or convolution, the computational metrics (element-wise or dot-product), and through full or structured transformation. Through the comparison, one can see that Score-CDM can be considered as a convolution model with a global attention score map, and its receptive field is flexible. For a clearer comparison among these methods, one can refer to the Appendix.

\subsection{Overview Diffusion Architecture}
We design a similar training process and backward process as PriSTI \cite{PriSTI}. 
In the training process, we train a map from diffusion step $t$ to noise $\epsilon$. In other words, our model learns to predict noise intensity in each diffusion step. We finally get a noise estimation function $\epsilon_{\theta}$ to denoise data step by step. To capture intervariate correlation, we additionally add an attention mechanism into $\epsilon_{\theta}$ to directly compute the data in variate dimensions. Its input is $Y$ and its output is the predicted noise.

\begin{algorithm}[!th]
\caption{Training process of Score-CDM.}
\label{alg:train}
\begin{algorithmic}[1]
\REQUIRE Incomplete MTS data ${X}$, the adjacency matrix $A$, the number of iterations $ite$, the number of diffusion steps $T$, noise levels sequence $\bar{\alpha}_t, \bar{\beta}_t \epsilon$.
\ENSURE Optimized noise prediction model $\epsilon_{\theta}$.
\STATE Insert the designed backbone into $\epsilon_{\theta}$ as the time exaction module;
\FOR{$i=1$ \TO $ite$}
    \STATE $\widetilde{X}^0 \gets \text{Mask}({X})$;
    \STATE $\mathcal{X} \gets \text{Interpolate}(\widetilde{X}^0)$;
    \STATE Sample $t \sim \text{Uniform}(\{1,\cdots,T\})$, $\epsilon\sim\mathcal{N}(0,\mathbf{I})$;
    \STATE Calculate noise  $\widetilde{X}^t \gets \bar{\alpha}_t \widetilde{X}^0 +  \bar{\beta}_t \epsilon$;
    \STATE Update the gradient $\nabla_{\theta}\left\Vert\epsilon-\epsilon_{\theta}(\widetilde{X}^{t}, \mathcal{X}, A, t)\right\Vert^2$. 
\ENDFOR
\end{algorithmic}
\end{algorithm}

\begin{algorithm}[!th]
\caption{Imputation process with Score-CDM.}
\label{alg:impute}
\begin{algorithmic}[1]
\REQUIRE The incomplete MTS data ${X}$, the adjacency matrix $A$, the number of diffusion steps $T$, the optimized noise prediction model $\epsilon_{\theta}$, and noise levels sequence $\bar{\alpha}_t, \bar{\beta}_t$.
\ENSURE  Missing values of the imputation target $\widetilde{X}^0$, where $\widetilde{X}$ is equal to $\widetilde{X}^0$.
\STATE $\mathcal{X} \gets \text{Interpolate}({X})$;
\STATE Set $\widetilde{X}^T\sim\mathcal{N}(0, \textbf{\text{I}})$;
\FOR{$t=T$ {\bfseries to} $1$}
    \STATE $\mu_{\theta}(\widetilde{X}^{t}, t) \gets \frac{1}{\sqrt{\bar{\alpha}_t}}\left(\widetilde{X}^{t}-\frac{\beta_t}{\sqrt{1-\bar{\alpha}_t}}\epsilon_{\theta}(\widetilde{X}^{t}, \mathcal{X}, A, t)\right)$;
    \STATE $\widetilde{X}^{t-1} \gets \mathcal{N}(\mu_{\theta}(\widetilde{X}^{t}, t), \sigma_{t}^{2}\bm{I})$;
\ENDFOR
\end{algorithmic}
\end{algorithm}


When using the trained noise prediction model $\epsilon_{\theta}$ for imputation, we aim to impute the incomplete MTS data ${X}$, and the interpolated conditional information $\mathcal{X}$ is constructed based on all observed values.  
The model receives $\widetilde{X}^T$ and $\mathcal{X}$ as inputs and generates samples of the imputation results through the reverse process in Eq. (\ref{eq:reverse_process}). 

\section{Experiment}

\begin{table*}[h]
\small
\centering
\renewcommand{\arraystretch}{1}

\setlength{\tabcolsep}{2.1mm}{
\begin{tabular}{ccccc|cccc|cccc|ccc}

\cmidrule[0.8pt]{1-15} & \multicolumn{2}{c}{}  
 & \multicolumn{4}{c|}{AQI-36}                                                                       & \multicolumn{4}{c|}{PEMS-BAY}                                                                 & \multicolumn{4}{c}{METR-LA}                                                \\ \cline{4-15} 
\multicolumn{3}{c}{Model}                      & \multicolumn{2}{c|}{$p\%$ = 25\%}              & \multicolumn{2}{c|}{$p\%$ = 50\%}                              & \multicolumn{2}{c|}{$p\%$ = 25\%}                       & \multicolumn{2}{c|}{$p\%$ = 50\%}                  & \multicolumn{2}{c|}{$p\%$ = 25\%}                       & \multicolumn{2}{c}{$p\%$ = 50\%} \\ 

\multicolumn{3}{c}{}                  & RMSE        & \multicolumn{1}{c|}{MAE} & RMSE                 & \multicolumn{1}{c|}{MAE}        & RMSE                 & \multicolumn{1}{c|}{MAE} & RMSE      & \multicolumn{1}{c|}{MAE}       & RMSE                 & \multicolumn{1}{c|}{MAE} & RMSE        & MAE        \\ \hline
\multicolumn{3}{c|}{Transformer}          & 29.46       & 16.26                    & 31.49                & \multicolumn{1}{c|}{17.45}      &  2.98                    & 1.63                         & 3.22          & \multicolumn{1}{c|}{1.74}          & 7.01                     & 2.82                         & 7.16        & 2.89       \\
\multicolumn{3}{c|}{BRITS}             & 28.76       & 15.72                    & 29.12                & \multicolumn{1}{c|}{16.01}      & 2.85                     & 1.59                         & 3.02          & \multicolumn{1}{c|}{1.67}          & 6.93                     & 2.80                         & 7.13        & 2.85       \\
\multicolumn{3}{c|}{SAITS}            & 29.85       & 16.24                    & 30.97                & \multicolumn{1}{c|}{17.18}      & 2.34                     & 1.35                         & 2.57          & \multicolumn{1}{c|}{1.42}          & 6.23                     & 2.66                         & 6.51        & 2.73       \\
\multicolumn{3}{c|}{CSDI}               & 14.52       & 7.71             & 16.93                & \multicolumn{1}{c|}{8.87} & 1.49                     & 0.76                         & 1.77          & \multicolumn{1}{c|}{0.85}          & 4.05                     & 2.27                         & 4.38        & 2.32       \\
\multicolumn{3}{c|}{TimesNet}            & 15.01       & 12.38                    & 17.49                & \multicolumn{1}{c|}{13.22}      & 1.88           & 0.92                         & 2.14 & \multicolumn{1}{c|}{0.97}          & 5.33                     & 2.49                         & 5.76        & 2.55       \\


\multicolumn{3}{c|}{GRIN}             & 12.93       & 7.93                     & 15.81   & \multicolumn{1}{c|}{9.02}           & 1.70 & 0.85     & 1.92      & \multicolumn{1}{c|}{0.91}      & 4.21 & 2.30    & 4.47        & 2.35       \\

\multicolumn{3}{c|}{SPIN}                 & 12.98 & 7.56                   & 16.53       & \multicolumn{1}{c|}{9.11}       & 1.62                     & 0.81                 & 1.83          & \multicolumn{1}{c|}{0.86} & 4.25           & 2.32                         & 4.51  & 2.39       \\
\multicolumn{3}{c|}{PriSTI}        & \underline{12.57}       & \underline{7.05}                     & \underline{14.68}                & \multicolumn{1}{c|}{\underline{8.25}}       & \underline{1.32}                     & \underline{0.71}                         & \underline{1.54}      & \multicolumn{1}{c|}{\underline{0.78}}      & \underline{3.84}                     & \underline{2.03}                         & \underline{4.16}        & \underline{2.09}       \\
\multicolumn{3}{c|}{\textbf{Score-CDM}}          & \textbf{12.14}            & \textbf{6.78}                         & \textbf{14.56}                     & \multicolumn{1}{c|}{\textbf{7.72}}           &  \textbf{1.21}                   & \textbf{0.65}                         & \textbf{1.33 }     & \multicolumn{1}{c|}{\textbf{0.69}}      & \textbf{3.59}                     & \textbf{1.93}                         & \textbf{3.85}        & \textbf{2.02}   \\
\hline
\end{tabular}}
\caption{Performance comparison of different methods on the three datasets in point missing scenario}
\label{tab:metric}
\end{table*}



\begin{table}[t]
 \small

\centering
\setlength{\tabcolsep}{4.5mm}{
\begin{tabular}{c | c  c  c  }
\cmidrule[0.4pt]{1-4}

\multicolumn{1}{c}{Model}&\multicolumn{1}{c}{METR} & \multicolumn{1}{c}{PEMS} & \multicolumn{1}{c}{AQI-36} \\
\cline{2-4}
\multicolumn{1}{c}{}&\multicolumn{3}{c}{Sensor failure probability 5\%} \\
\hline
BRITS & 5.87  & 4.14   & 24.09  \\
SAITS & 4.73 & 3.88   & 20.78  \\
Transformer & 6.03  & 3.69  & 29.21  \\
GRIN & 3.05  & 2.26   & 15.62  \\
SPIN & 2.74 & 1.78  & 14.29  \\
PriSTI & \underline{2.70} & \underline{1.66}  & \underline{14.01}  \\
\hline
\textbf{Score-CDM} & \textbf{2.60} & \textbf{1.55} & \textbf{13.74} \\
\hline
\end{tabular}}
\caption{MAE comparison of different methods with a sensor failures probability $q\%$ in block missing scenario}
\label{tab:result_block}
\end{table}


\begin{table}[t]
\small
\renewcommand{\arraystretch}{1}

\centering
\setlength{\tabcolsep}{4mm}{
\begin{tabular}{l | r  r | r  r }
\cmidrule[0.4pt]{1-5}

\multicolumn{1}{c}{Model}&\multicolumn{2}{c}{METR-LA} & \multicolumn{2}{c}{PEMS-BAY} \\
\cline{2-5}
\multicolumn{1}{c}{}  & \multicolumn{1}{c|}{75 \%} & \multicolumn{1}{c|}{95 \%} &  \multicolumn{1}{c|}{75 \%} & \multicolumn{1}{c}{95 \%} \\
\hline
 BRITS & 3.02  & 5.19  & 2.17  & 3.91 \\
 SAITS & 3.74  & 6.72  & 2.96  & 7.40  \\
 Transformer & 2.71  & 5.13  & 1.13  & 2.70   \\
 GRIN  & 2.39  & 4.08  & 1.09  & 2.70   \\
 SPIN &  2.24  & 2.89  & 1.09  & \underline{2.26} \\
 PriSTI & \underline{2.21}  & \underline{2.89}  & \underline{1.08}  & 2.27 \\
\hline
\textbf{Score-CDM} & \textbf{2.14} & \textbf{2.86} & \textbf{1.06} & \textbf{2.23} \\
\hline
\end{tabular}}
\caption{MAE comparison of different methods with high data missing percentage (75\% and 95\%) in the point missing scenario}
\label{tab:result_sparse}
\end{table}




\subsection{Experiment Setup}\label{sec:exp_set}
\textbf{Dataset.} We evaluate the performance of our model on three real-world datasets METR-LA, AQI-36, and PEMS-BAY that are widely adopted for MTS imputation in previous works. METR-LA and PEMS-BAY are the traffic flow datasets collected from traffic sensors in Los Angeles County Highway and Bay Areas in California. AQI-36 is collected from 36 AQI sensors distributed across the city of Beijing. The detailed dataset statistics are given in the appendix.
For the dataset AQI-36, we adopt the same evaluation strategy as the previous work \cite{yi2016st}.
For the traffic datasets METR-LA and PEMS-BAY, more details will show in Appendix.

\textbf{Baselines.} We compare Score-CDM with the following baselines. 
\begin{itemize}
    \item \textbf{Transformer} \cite{attention} is based on multi-head self-attention mechanism.  
    \item \textbf{BRITS} \cite{BRITS} employs bidirectional RNN and MLP to learn spatio-temporal information for MTS imputation. 
    \item \textbf{CSDI} \cite{csdi} is a recent SOTA MTS imputation method that is based on the conditional diffusion probability model. 
    \item \textbf{TimesNet} \cite{timesnet} contains a self-organized convolution model for time series imputation. 
    \item \textbf{SPIN} \cite{spin} employs threshold graph attention and temporal attention to jointly model the spatio-temporal dependencies of time series. 
    \item \textbf{GRIN} \cite{grin} is a bidirectional GRU-based method with a graph neural network. 
    \item \textbf{SAITS} \cite{saits} is based on diagonally-masked self-attention mechanism for MTS imputation. 
    \item \textbf{PriSTI} \cite{PriSTI} incorporates spatial conditions into attention to reduce the discrepancy between the missing time series values and the ground truth. 
\end{itemize}

To study whether each module of Score-CDM is useful, we also compare Score-CDM with the following two variants. 
\begin{itemize}
\item \textbf{w/o(S2TWB)}: This variant of Score-CDM removes the S2TWB block. 
\item \textbf{w/o(SCM)}: This variant of Score-CDM removes the score-weighted convolution module SCM. 
\end{itemize}

In the experiment, the baselines Transformer, SAITS, BRITS, GRIN, and SPIN are implemented by the code\footnote{https://github.com/Graph-Machine-Learning-Group/spin} provided by the work \cite{spin}. 

\textbf{Evaluation metrics.} We apply Mean Absolute Error (MAE) and Root Mean Squared Error (RMSE) defined as follows to evaluate the model performance.
\[
MAE = \frac{1}{T} \sum_{k=0}^{L} \left\| {X}_{k} - \widetilde{X}_{k} \right\| 
\]
\[
RMSE = \sqrt{\frac{1}{T} \sum_{k=0}^{L} ( \left\| {X}_{k} - \widetilde{X}_{k} \right\|^2_{F} }
\]
where ${X}_{t}$ is the imputed time series at time $t$ and $\widetilde{X}_{t}$ is the corresponding ground truth.




\begin{figure}[!t]
	\centering

	\subfloat[METR-LA (MAE)\label{fig:wiki_cs}]{\includegraphics[width=.24\textwidth]{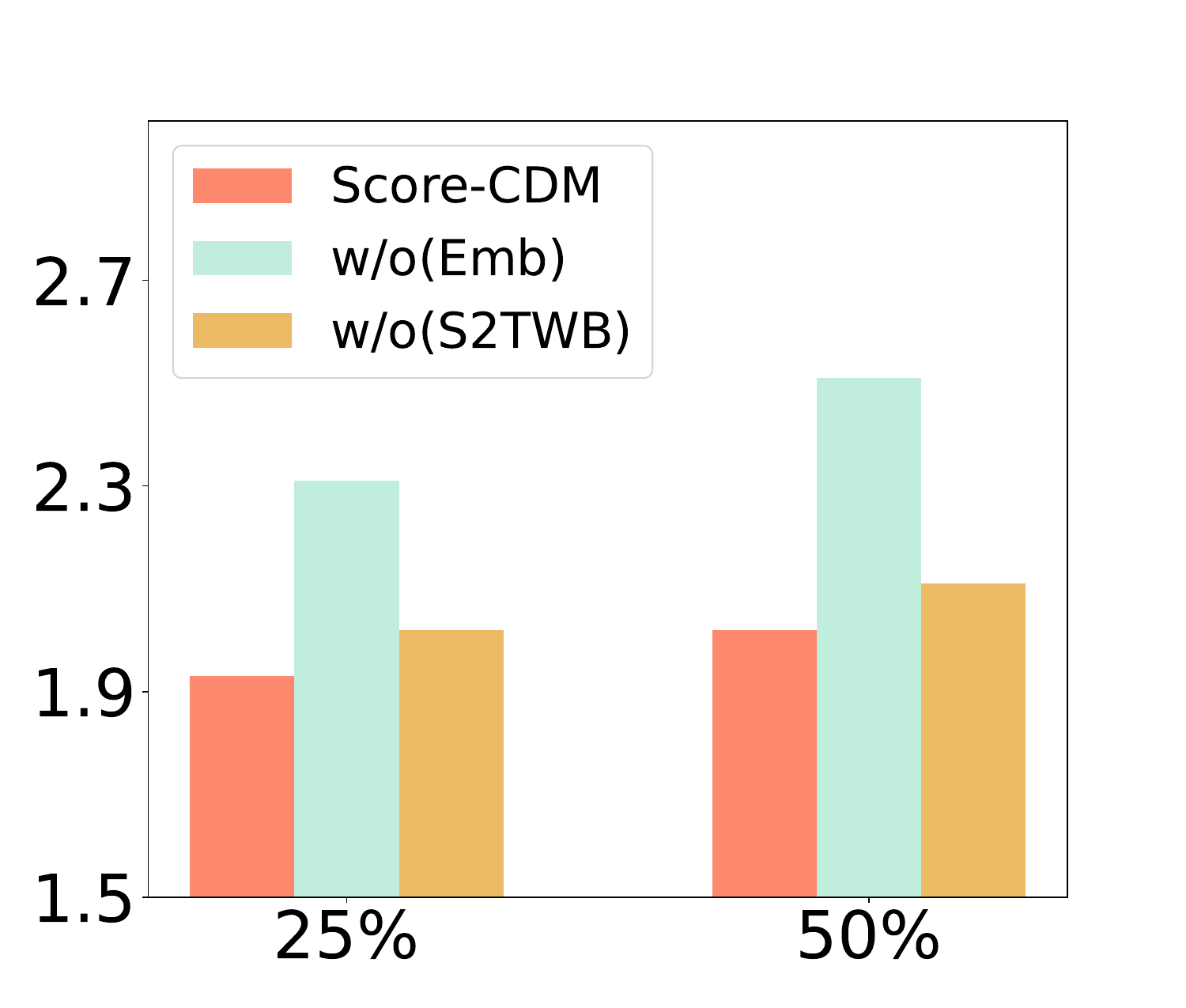}}
	\subfloat[PEMS-BAY (MAE)\label{fig:amazon_cs}]{\includegraphics[width=.240\textwidth]{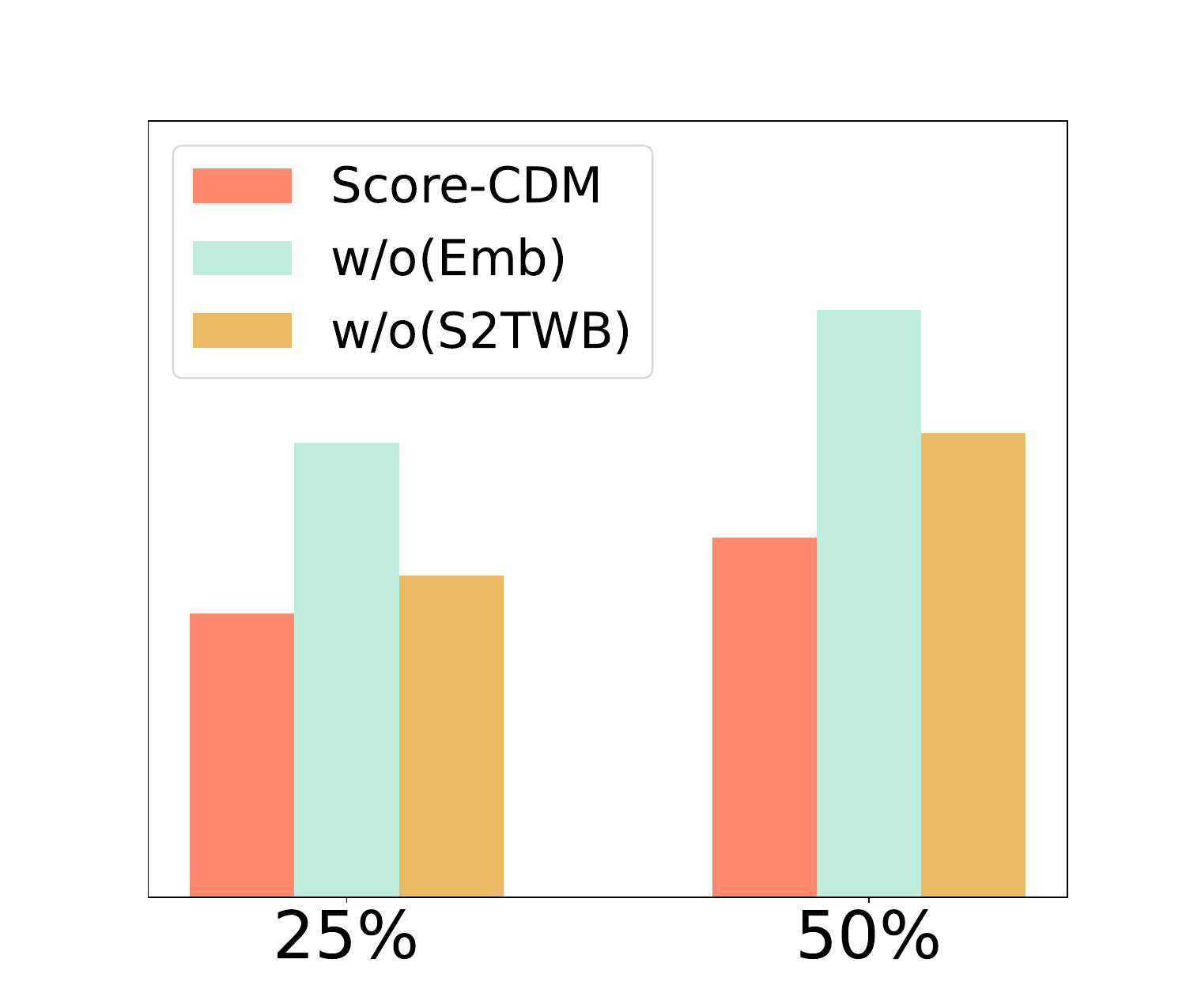} }\quad
	\subfloat[METR-LA (RMSE)\label{fig:amazon_photo}]{\includegraphics[width=.240\textwidth]{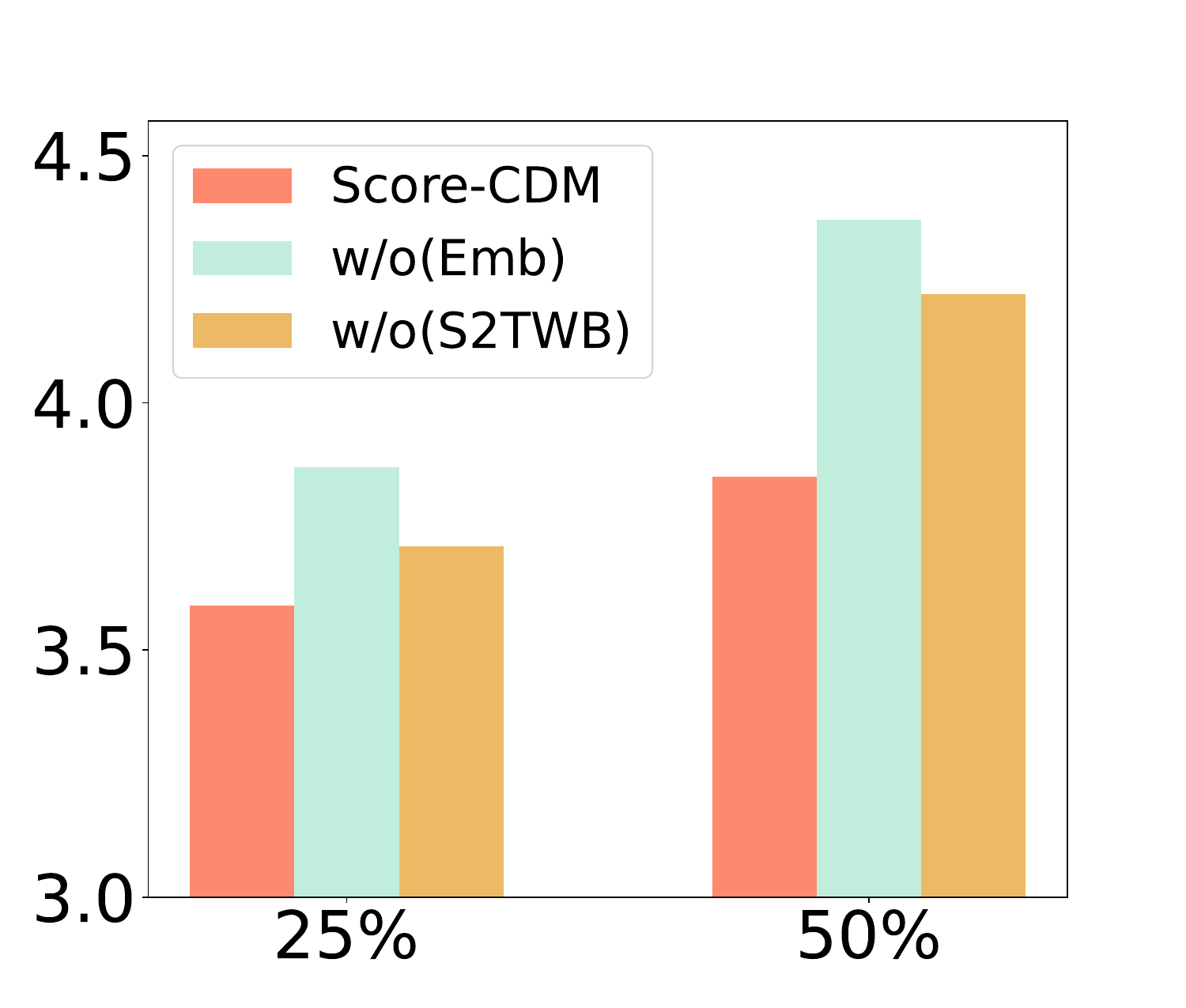} }
	\subfloat[PEMS-BAY (RMSE)\label{fig:coauthor_cs}]{\includegraphics[width=.240\textwidth]{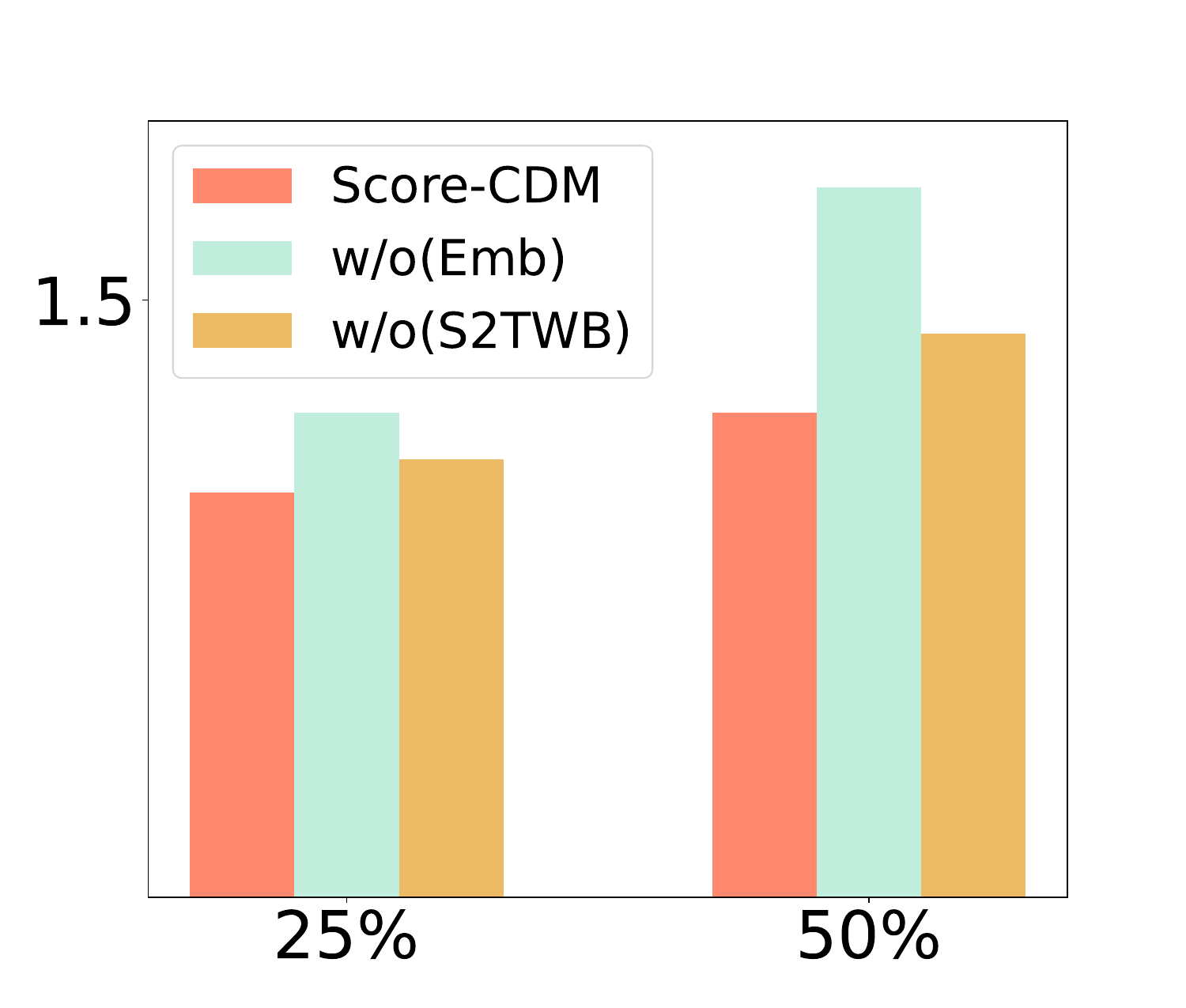} }
     \caption{Performance comparison between Score-CDM and two variant models on the point data missing scenarios}
	\label{fig:abla}
\end{figure}

\subsection{Experiment Result}

We first compare the performance of different methods in the point missing scenario with the data missing rate $p\%$ setting to $25\%$ and $50\%$, respectively. The experiment result is shown in Table \ref{tab:metric}. The best result is highlighted in bold font, and the second-best result is underlined. From Table \ref{tab:metric}, one can observe that Score-CDM consistently outperforms all the baseline methods in both cases and over all the datasets. Specifically, Score-CDM improves the performance of the best baseline PriSTI by $3\%$-$5\%$ in terms of MAE on AQI-36, by $9\%$-$12\%$ on PEMS-BAY dataset, and by $5\%$-$7\%$ on METR-LA dataset. This demonstrates that Score-CDM can effectively balance the local and the global temporal information of time series. Compared with attention-based diffusion models CSDI and PriSTI, Score-CDM performs better in all the datasets, verifying the effectiveness of the extracted score map by SCM and the self-attention kernel of S2TWB to construct the flexible receptive field. Compared with RNN-based methods BRITS, the performance improvement of Score-CDM is much more significant. For example, the RMSE of BRITS on AQI-36 is 28.76 when $p\%=25\%$, while the RMSE of Score-CDM is dropped to only 12.14. GRIN is also an RNN model, but its performance is much better than BRITS by incorporating GNN models. One can see that SPIN performs best among all the attention-based methods, indicating that integration of both temporal and spatial information can significantly enhance model performance for MTS imputation. However, the performance of SPIN is still inferior to PriSTI, which suggests that diffusion models is truly powerful in MTS imputation due to their strong generative capability. TimesNet's performance is moderate among all the methods. This is because its receptive field is smaller than attention methods, and thus is less effective to capture long-term temporal features in time series data.

For the block missing scenario, we set the sensor missing probability $q\%=5\%$ to mimic that $5\%$ sensors fail in 1 to 4 hours without any time series data observations. We compare Score-CDM with six strong baselines. The result is shown in Table \ref{tab:result_block}. It shows that Score-CDM still outperforms the baseline methods on the three datasets, demonstrating its superior performance in the block data missing scenario. PriSTI achieves the best performance among all the baselines, but it is still inferior to Score-CDM. For example, Score-CDM outperforms PriSTI by more than 3\% in terms of MAE on METR-LA \& AQI-36, and by more than 12\% on PEMS-BAY. To further evaluate the performance of different methods under very high point data missing percentages, we compare Score-CDM against the baselines when $p\%=75\%$ and $p\%=95\%$. The result is shown in Table \ref{tab:result_sparse}. It demonstrates again that Score-CDM outperforms all the baselines when the available time series observations are very sparse.

\subsection{Ablation Study}


To examine whether the designed two modules SCM and S2TWB work, we conduct the ablation study to compare Score-CDM with its two variant models w/o[S2TWB] and w/o[SCM]. Figure \ref{fig:abla} shows the result. One can see both modules are useful to the model, as removing each one of them will lead to remarkably performance drop in all four cases. One can also see that SCM has a larger impact on the model performance compared with S2TWB, because removing it leads to a more significant performance decline. This implies that the global temporal features are critical to MTS imputation and the proposed SCM can effectively capture the global features. When 25\% point data are missing on the PEMS-BAY dataset, the performance of w/o[S2TWB] drops by over 5\% compared with Score-CDM in terms of MAE, and the performance drop is up to 8.7\% when the 50\% data are missing, which verifies S2TWB is also important to the performance improvement. This indicates a pronounced periodicity of the traffic flow time series in the PEMS-BAY dataset, characterized by a prevalence of local temporal features. The proposed S2TWB in Score-CDM can effectively capture this periodicity by extracting the corresponding frequencies in the spectral domain. 

\begin{figure}[t]
	\centering
	\includegraphics[width=1\linewidth]{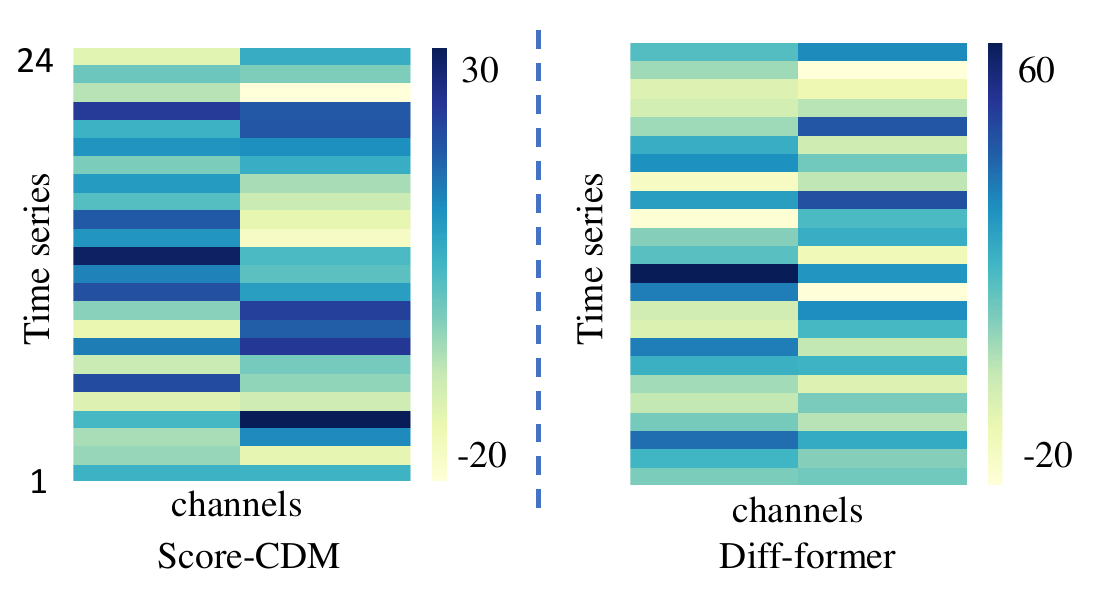}
	\vspace{-0.5mm}
	\caption{A case study to visualize the score maps of Score-CDM and Diff-former.}
	\vspace{-0.5mm}
	\label{fig:score}
\end{figure}

\begin{figure}[!t]
	\centering
	\includegraphics[width=1\linewidth]{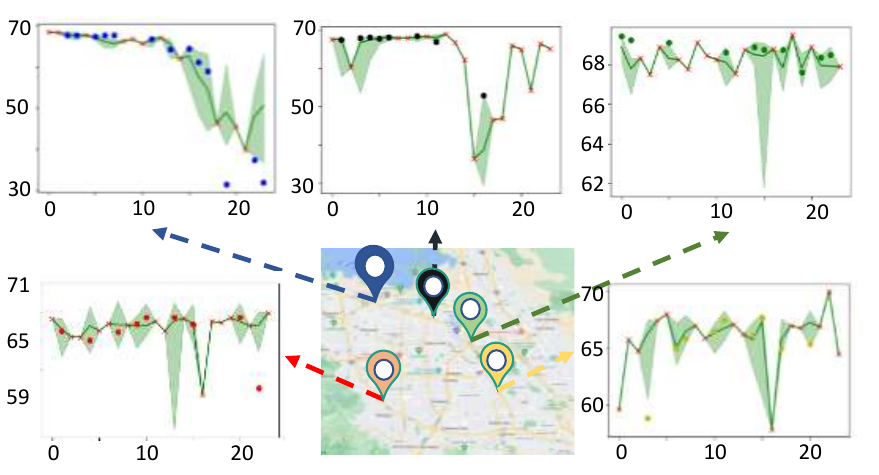}
	\vspace{-1mm}
	\caption{A case study to show the traffic flow time series data imputation results on 5 road sensors in METR-LA dataset.}
	\vspace{-2mm}
	\label{fig:METR-LA}
\end{figure}

\subsection{Case Study}
To further show the effectiveness of Score-CDM, we give a case study to visualize the learned score map in Figure \ref{fig:score}. To make a comparison, we also present the score map learned by Diff-former. Diff-former applies the attention mechanism as the denoising function of the diffusion model to learn the score map and extract time features. We select a traffic flow time series whose length is 24 and with two channels from METR-LA. The darker color in the figure represents a higher score, while the light color represents a smaller score. It shows that Score-CDM can better capture both local and global temporal features compared with Diff-former as the high attention scores are distributed over different locations on the score map, while the high score of Diff-former only located at one or two areas in its score map.  

In Figure \ref{fig:METR-LA}, we give a case study to show the traffic flow time series imputation results of 5 road sensors by Score-CDM in METR-LA dataset. Each subfigure represents the imputation result of a sensor, and the time windows of all sensors are in one day (24 hours). The red crosses represent observations, and dots of various colors represent the ground truth of the missing values. The solid green line is the imputation result by Score-CDM, and the green shadow represents the quantile between 0.05 to 0.95. One can see that the imputed values denoted by the green curves are very close to the ground truth missing time series points and the observations, which demonstrates the desirable time series data imputation performance of Score-CDM.

\section{Conclusion}

This paper proposed a Denoising Diffusion Probabilistic Model based method Score-CDM for multivariate time series imputation. Significantly different from existing methods, Score-CDM employed the specially designed denoising function to adaptively capture and balance the global and local temporal features in time series. The designed denoising function contained two modules, SCM and ARM. SCM generated a score map that contained a global weighting of the entire time series, and ARM module adapted S2TWB to generate a flexible receptive field of the score map. Extensive evaluations over three real-world datasets showed the effectiveness of the proposed model.

\appendix

\section{Complexity Analysis.}  As shown in Table \ref{tab:comparison1}, we compare our method with five transformer models including Transformer, Reformer \cite{reformer}, Linear Transformer \cite{lintransformers}, Performer \cite{performer}, and Attention Free Transformer in terms of time complexity and space complexity. The comparison shows the efficiency of Score-CDM and its variant.

\begin{table}[!h]

\small
   
\vspace{-0.25em}
    
    \setlength{\tabcolsep}{0.30mm}{
    \centering
    \begin{tabular}{ccc}
    \cmidrule[0.8pt]{1-3}
Model & Time & Space \\
\hline
Transformer &$O(T^2d)$&$O(T^2 + Td)$ \\
Reformer &$O(T\log{T}  d)$&$O(T\log{T} + Td)$\\
Linear Transformer &$O(T d^2)$&$O(Td + d^2)$ \\
Performer &$O(T d^2\log{d})$&$O(Td\log{d} + d^2\log{d})$\\
AFT &$O(T^2d)$&$O(Td)$\\
\textbf{Score-CDM}&$O(T d + T log T)$&$O(Td)$\\
(w/o[S2TWB])&$O(T d)$&$O(Td)$\\
  \hline
\end{tabular}
}
 \caption{Complexity comparison with different Transformers. $T$ and $d$ denote the sequence length and channel dimension, respectively.}
\vspace{-0.25em}
\label{tab:comparison1}
\end{table}

\section{Comparison of Different Models}
We compare Score-CDM with four classic models including Attention Free Transformer, Transformer, TCN, and Dlinear (MLP-based Model) from four aspects, whether can be coded as attention or convolution, the computational metrics (element-wise or dot-product), and through full or structured receptive field. Through the comparison, one can see that Score-CDM can be considered a convolution model with a global attention score map, and its receptive field is flexible. For a clearer comparison among these methods, one can refer to Table \ref{tab:model compare}.
\begin{table}[!h]
\small

\vspace{-0.25em}
    
    \centering
    \begin{tabular}{ccccc}
\cmidrule[0.8pt]{1-5}
Model & Attent & Field& Comput & Conv \\
\hline
AFT  &   \XSolidBrush        & full                & element & \XSolidBrush \\
Transformer & \checkmark & full                & dot  & \XSolidBrush \\
TCN & \XSolidBrush & \textbf{local \& structured}                & element  & \checkmark \\
DLinear & \XSolidBrush          & full                & dot  & \XSolidBrush \\

Score-CDM& \checkmark & \textbf{full \& structured} & element & \checkmark\\

  \hline
\end{tabular}
\caption{Comparison of different models. In this table, we shorthand Attention as Attent, receptive field as Field, Computation as Comput, and Convolution as Conv.}
\label{tab:model compare}
\vspace{-0.25em}

\end{table}{}

As the table shows, the receptive field of TCN is structured but local. Compared to TCN, the receptive field of Score-CDM is structured and global.

\section{Convolution Theorem}
To use the Fast Fourier Transform, we need to introduce the convolution theorem first, which is a communication theory as follows,

$$\mathcal{F}(K \ast X) = \mathcal{F}(K) \cdot \mathcal{F}(X)
$$
\newline
where $\ast$ is a convolution operation, $\cdot$ is a multiply operation. $\mathcal{F}$ represents Fast Fourier Transform, which can be used to convolve time series. For time series $X$ and convolution kernel $K$, they can be computed by using FFT which is the same as a time convolution operation.

\section{Dataset}

For the traffic datasets METR-LA and PEMS-BAY, we artificially inject some missing values by following \cite{grin} to construct the imcomplete data. We evaluate the model on two data missing scenarios, block missing and point missing. In the block missing scenario, we first randomly mask 5\% of the time series data, and then for each sensor we mask its data ranging from 1 to 4 hours with a probability $q\%$ as in \cite{spin} to mimic sensor failure. For the point missing case, we randomly mask $p\%$ of all the time series observations. 

Additionally, as shown in Table \ref{tab:dataset compare}, We evaluate the performance of our model on three spatial time series datasets, METR-LA, AQI-36, and PEMS-BAY. METR-LA is a dataset used in traffic flow prediction and imputation. It contains 207 traffic sensor nodes in Los Angeles County Highway with a minute-level sampling rate. AQI-36 is collected from 36 AQI sensors distributed across the city of Beijing. This dataset serves as a widely recognized benchmark for imputation techniques and includes a mask used for evaluation that simulates the distribution of actual missing data \cite{yi2016st}. For a specific month, such as January, this mask replicates the patterns of missing values from the preceding month. Across all scenarios, the valid observations that have been masked out are employed as targets for evaluation. PEMS-BAY is an open dataset used for traffic flow prediction and analysis, primarily covering the transportation network of the Bay Area in California, USA. The dataset comprises 325 sensor nodes with a sampling interval of 5 minutes, and it contains a total of 16,937,700 data points. 

\begin{table}[!h]
\small
    
\vspace{-0.25em}
    
    \centering
    \begin{tabular}{ccc}
\cmidrule[0.8pt]{1-3}
Dataset & Node & Time step \\
\hline
METR-LA  & 207 & 34272\\
PEMS-BAY & 325 & 52116\\
AQI-36 & 36 & 52116\\

\hline
\end{tabular}
\caption{Comparison of different datasets.}
\label{tab:dataset compare}
\vspace{-0.25em}
\end{table}{}
In total, each dataset will be artificially masked 25\% or 50\% values at random. For the two datasets METR-LA and PEMSBAY, we partition the entire data into training, validation, and testing sets by a ratio of 8 : 1: 1. We evaluate our model performance under two metrics Mean Absolute Error (MAE) and Root Mean Square Error (RMSE).

\section{Experiment Settings}

\begin{table}[!h]
  \centering
  
  \setlength{\tabcolsep}{1mm}
  \resizebox{0.9\columnwidth}{!}{
    \begin{tabular}{cccc}
    \toprule
    Description & AQI-36 & METR-LA& PEMS-BAY\cr
    \midrule
    Batch size  & 16 & 16 & 16 \cr
    Time length $L$ & 24 & 24 & 24 \cr
    Epochs & 200 & 200 & 200 \cr
    Learning rate & 0.001 & 0.001 & 0.001 \cr
    Channel size $d$ & 64 & 64 & 64 \cr
    Minimum noise level $\beta_1$ & 0.0001 & 0.0001 & 0.0001 \cr
    Maximum noise level $\beta_T$ & 0.5 & 0.2 & 0.2 \cr
    Diffusion steps $T$ & 50 & 50 & 50 \cr
    \bottomrule
    \end{tabular}}
    \caption{The hyperparameters of Score-CDM for all datasets.}
  \label{tab:exp_setting}
\end{table}

For the hyperparameters of Score-CDM, the batch size is 16. The hyperparameter for diffusion model includes a minimum noise level $\beta_1$ and a maximum noise level $\beta_T$. We adopted the quadratic schedule for other noise levels following \cite{csdi}, which is formalized as:
\begin{equation}
    \beta_t=\left(\frac{T-t}{T-1}\sqrt{\beta_1}+\frac{t-1}{T-1}\sqrt{\beta_T}\right)^2.
\end{equation}
We summarize the hyperparameters of Score-CDM in Table \ref{tab:exp_setting}.

\section{Discussion}

As mentioned in Section 4.1, Eq.\ref{eq:element-mixing} shows the additional part, which we call channel mixing, compared to the attention-free transformer like AFT, which leads to enough message passing through time and channel dimension on element view.

To address the challenge of how to reduce complexity with a comparable result, recent works like RWKV \cite{rwkv} have designed an additional module named channel mixing to fill the missing part which is serial to time mixing.

For our work, the S2TWB \& ARM can be regarded as a channel-mixing module, which is paralleled with SCM.  The whole design ensures a comparable result in evaluation.

\section*{ACKNOWLEDGMENTS}
This research was funded by the National Science Foundation of China (No. 62172443 and 62206303), Hunan Provincial Natural Science Foundation of China (No. 2022JJ30053) and Science and Technology Innovation Program of Hunan Province(No.2023RC3009).

\section*{CONTRIBUTIONS}
Shunyang Zhang and Senzhang Wang are the co-first authors who contribute equally. Jian Zhang is the Corresponding author.

\bibliographystyle{named}
\bibliography{ijcai24}

\end{document}